# RFCBF: enhance the performance and stability of Fast Correlation-Based Filter

Xiongshi Deng, Min Li, Lei Wang, Qikang Wan

*Abstract*—Feature selection is a preprocessing step which plays a crucial role in the domain of machine learning and data mining. Feature selection methods have been shown to be effective in removing redundant and irrelevant features, improving the learning algorithm's prediction performance. Among the various methods of feature selection based on redundancy, the fast correlation-based filter (FCBF) is one of the most effective. In this paper, we proposed a novel extension of FCBF, called RFCBF, which combines resampling technique to improve classification accuracy. We performed comprehensive experiments to compare the RFCBF with other state-of-the-art feature selection methods using the KNN classifier on 12 publicly available data sets. The experimental results show that the RFCBF algorithm yields significantly better results than previous state-of-the-art methods in terms of classification accuracy and runtime.

## I. INTRODUCTION

FEATURE selection can not only remove the irrelevant and redundant feature information in the data, but also reduce the data dimension, noise interference, and algorithm complexity. In this way, the model will become simple and easy to understand, the performance of data mining will be improved, and clean and understandable data can be prepared for later predictive analysis. In the field of data mining, feature selection has become a research hotspot. [1], [2].

At present, feature selection methods can be categorized into two main approaches based on their interaction with the subsequent learning algorithms, namely, filter method, and wrapper method [3]–[5]. The filter method generally uses distance, information, dependency, and consistent evaluation criteria to enhance the correlation between features and classes, weaken the correlation between features. Thereby, a subset of features that can better represent the characteristics of the original data can be selected [6]–[8]. This type of method has high efficiency in feature selection, but it is sensitive to noise data. In practical application, this type of method is generally used for preliminary feature selection. The wrapper method is strongly associated with the classifier used. This method trains the classifier directly using the selected feature subset in the feature screening process and then evaluates the selected features by the performance of the classifier on the validation set. Because it directly uses the learning algorithm to evaluate feature subsets, which is conducive to the extraction of key features. Therefore, the obtained feature subsets have higher classification performance and higher prediction accuracy. However, the wrapper method has the problem of high time complexity and is not appropriate for super large Large-scale data mining tasks. Compared with the wrapper method, the time complexity of the filter method is much lower.

In classification tasks, feature selection can help to improve the prediction accuracy by removing the noisy features and avoiding overfitting. But feature selection can also be very challenging, especially when there is a large number of features (high-dimension) and few training samples. For this reason, the filter method is the most used. For instance, the most typically used filter methods include chi-square statistics, information gain, Relief, Fast Correlation-Based Filter (FCBF), MRMR (Maximum Relevance, Minimum Redundancy), and correlation measurement [8]–[12].

Among the above methods, FCBF is one of the most efficient methods. At present, there are many kinds of research on the extension of FCBF. For instance, considering that FCBF cannot select a specific number of features, Senliol et al. used different search strategies to enhance FCBF and proposed an FCBF extension, called FCBF# [13]. Experimental results show that the proposed FCBF# algorithm can obtain higher classification accuracy. In [14], Egea et al. proposed an approach called FCBF in Pieces (FCBFiP), which allows users to control both the algorithm computing time and the intercorrelation among the features contained in the resulting subset. Aiming at the size of the selected feature subset is not considered by FCBF, and weakly relevant features are too inclined to be eliminated. In [2], Zeng et al. proposed a new approximate Markov blanket definition for FCBF, which proposed algorithm can dynamically adjust the number of selected features, and the performance is better than FCBF.

This paper discusses a novel extension of FCBF, called RFCBF, which combines resampling technique to improve classification accuracy. Different from FCBF using SU (Symmetrical Uncertainty) to calculate the overall feature score, the proposed RFCBF algorithm is iteratively randomly sampling part of the feature values for calculation, which eliminates the impact of the overall feature values. Moreover, RFCBF performs multiple SU calculations, which can avoid using SU calculation once to remove features with low ratings but strong relevance. For a detailed introduction, see Algorithm 2 in Section II.

The remainder of this paper is organized as follows. In section II, we review the FCBF algorithm and present our proposed algorithm in detail. Next, in section III, we described the basic requirements and settings of the experiment. In section IV, we compare the RFCBF algorithm with three other state-of-the-art feature selection algorithms on 12 publicly available data sets. Finally, in section V, we present the conclusions and future work.

## II. METHODOLOGY

In this section, an improved feature selection algorithm named RFCBF is introduced. The algorithm is based on FCBF extension and improvement. This section first provides a brief description of FCBF, then presents the procedure of RFCBF in detail.

### A. FCBF

FCBF algorithm (full name Fast Correlation-Based Filter Solution), is a fast-filtering feature selection algorithm. It was first proposed by Yu et al. in 2004, using a backward sequential search strategy to quickly and effectively find the optimal feature subset [15]. FCBF is an algorithm based on SU. If the information gain is used to select features directly, the feature with the larger information gain value will be selected. The symmetric uncertainty corrects the bias of using information gain to select features and normalizes the information gain. This makes it relatively fair in the comparison of feature relevance. Hence, the authors propose to use SU instead of IG (Information Gain) as a measure of whether a feature is related to category C or whether it is redundant. Algorithm 1 shows an overview of the FCBF algorithm in detail.

The SU is based on the concepts of entropy and conditional entropy to measure the correlation between features. It is defined as follows:

$$SU(X, Y) = 2\left[\frac{IG(X,Y)}{E(X)+E(Y)}\right] \qquad (1)$$

The IG is defined as:

$$IG(X, Y) = E(X) - E(X|Y) \qquad (2)$$

where $E(X)$ presents the entropy of a feature $X$, corresponds to the amount of information contained in or supplied by a source of information, is defined as:

$$E(X) = -\sum_{i=1}^{c} P(x_i) * log_2\left(P(x_i)\right) \qquad (3)$$

where $P(x_i)$ is the probability of $X$ to take $x_i$, and $c$ is the number of categories. The entropy of $X$ after observing values of another feature $Y$ is defined as:

$$E(X|Y) = \sum_{i=1}^{c_y} P(y_i) \sum_{j=1}^{c} P(x_j|y_i) log_2(P(x_j|y_i)) \qquad (4)$$

It can be seen from equation (1) that the $SU$ is a form of normalization of $IG$. When the value of $SU=1$, it means that X and Y are completely correlated, that is, from $X \rightarrow Y$, or $Y \rightarrow X$. Meanwhile, when the value of $SU=0$, X and Y can be obtained independently.

---

**Algorithm 1: FCBF**

---
**input:** $S(F_1, F_2, \ldots, F_N, C)$  // a training data set
      $\delta$        // a predefined threshold
output: $S_{best}$  // selected feature subset
1 **begin**
2   for $i = 1$ to N do begin
3     calculate $SU_{i,c}$ for $F_i$;
4     if ($SU_{i,c} \geq \delta$)
5       append $F_i$ to $S'_{list}$;
6   end;
7   order $S'_{list}$ in descending $SU_{i,c}$ value;
8   $F_p = getFirstElement(S'_{list})$;
9   do begin
10     $F_q = getFirstElement(S'_{list}, F_p)$;
11     if ($F_q <> NULL$)
12       do begin
13       $F'_q = F_q$;
14       if ($SU_{p,q} \geq SU_{q,c}$)
15         remove $F_q$ from $S'_{list}$;
16       $F_q = getNextElement(S'_{list}, F'_q)$;
17     end until ($F_q == NULL$);
18     $F_p = getNextElement(S'_{list}, F_p)$;
19   end until ($F_p == NULL$);
20 $S_{best} = S'_{best}$;
21 end;

---

### B. RFCBF Algorithm

Previous experiments [12] show that FCBF is an efficient and fast algorithm that uses interdependence of features together with the dependence on the class. FCBF selects a predominant feature, and then removes some redundant features based on the predominant feature. On account of FCBF brings the overall data into the calculation of the score of each feature on the basis of the SU, and then removes redundant features. This can lead to the removal of features with low ratings but strong relevance. The impact of the SU on the calculation of the whole feature values is not considered. To address this issue, we propose a feature selection algorithm that combines FCBF and sample resampling technology, called RFCBF. The proposed RFCBF algorithm does not calculate the SU value of the feature as a whole, but randomly samples a part of the data from the data set and performs multiple resampling to eliminate the impact of the feature value on the SU calculation. Compared with the FCBF algorithm, the RFCBF is more robust and can avoid overfitting. Algorithm 2 shows an overview of the RFCBF algorithm in detail.

In Algorithm 2, $sample()$ is the resampling technique that we incorporate into the SU calculation. We use a simple random sampling method to repeatedly extract part of the sample data from the training sample with a certain sampling probability, and then to calculate the SU value of each feature. This can improve the reliability of the SU value of each feature calculated by the sample.

In Section A, we briefly introduced the FCBF algorithm. The overall complexity of the FCBF is $O(MNlogN)$, where $M$ is the number of samples and $N$ is the number of features. For the proposed RFCBF algorithm, due to the integration of resampling technology, the time complexity for the first part is $O(TN)$, where $T$ is the number of sampling times. The time complexity for the second part is $O(TN^2 logN)$. Since the calculation of $SU$ for a pair of features is linear in term of the number of instances $M$ in a data set. Therefore, the overall time complexity of the RFCBF is $O(GTMN^2 logN)$, where $G$ is the sampling probability.

---

**Algorithm 2: RFCBF**

**input:** $S(F_1, F_2, \ldots, F_N, C)$ // a training data set
  $\delta$  // a predefined threshold
  $T$, $G$ // sampling times, sampling probability, respectively
**output:** $S_{best}$ // selected feature subset
1 **begin**
2 for $m = 1$ to $T$ do begin
3   $S_m = sample(S, G)$; // random resampling
4   for $i = 1$ to $N$ do begin
5     calculate $SU_{i;c}$ for $F_i$
6     append $SU_{i;c}$ to $SU_{m;i}$;
7   end;
8   $SU_{i;c} = sum(SU_{m;i})/T$;
9   if $(SU_{i,c} \geq \delta)$
10    append $F_i$ to $S'_{list}$;
11 end;
12 order $S'_{list}$ in descending $SU_{i,c}$ value;
13 $F_p = getFirstElement(S'_{list})$;
14 do begin
15   $F_q = getFirstElement(S'_{list}, F_p)$;
16   if $(F_q <> NULL)$
17     do begin
18       $F'_q = F_q$;
19       repeat 2-7 to calculate $SU_{p,q}$
20       if $(SU_{p,q} \geq SU_{q,c})$
21         remove $F_q$ from $S'_{list}$;
22         $F_q = getNextElement(S'_{list}, F'_q)$;
23       else $F_q = getNextElement(S'_{list}, F_q)$;
24     end until $(F_q == NULL)$;
25   $F_p = getNextElement(S'_{list}, F_p)$;
26 end until $(F_p == NULL)$;
27 $S_{best} = S'_{best}$;
28 end;

---

Compared with FCBF, which has only one parameter threshold $\delta$, the proposed RFCBF involves three parameters, namely threshold $\delta$, sampling times, and sampling probability, respectively. In the experiment, to ensure that each sample has the same probability of being selected, we set the sampling probability value to 0.5.

## III. EXPERIMENT SETTIN

We conducted experiments to verify the performance of the proposed algorithm. The data sets required for the experiment, classification algorithm, and performance measure are presented in this section. The results were generated on a PC equipped with a Core i5-5200U CPU and 12 G of memory.

### A. Data Sets

In the experiment, we used 12 data sets to verify the performance of the proposed algorithm. Among the 12 data sets, 6 data sets are from the UCI repository [16], and the rest are from KEEL [17]. Table 1 outlines the characteristics of the 12 data sets, including the numbers of instances, features, and classes. Among these data sets, 3 data sets were multi-class, whereas the others were binary.

As some of the originally obtained data sets contained missing values, we performed missing value interpolation to process the missing values. In our experiments, we performed the k-nearest-neighbor interpolation to replace the missing values. Each sample's missing values were imputed using the mean value from n nearest neighbors in the training set. After the missing values were processed, we normalize the feature values of the data so that the processed data conforms to the standard normal distribution, that is, the mean is 0 and the standard deviation is 1, and the conversion function is:

$$x^* = \frac{x - \mu}{\sigma} \qquad (5)$$

where $\mu$ is the mean of all sample data, and $\sigma$ is the standard deviation of all sample data.

TABLE 1 SUMMARY OF DATA SETS

| Data set | Instance | Feature | Class |
|---|---|---|---|
| clean1 | 476 | 166 | 2 |
| Hill_Valle | 606 | 100 | 2 |
| ionosphere-10an-nn | 351 | 33 | 2 |
| Libras Movement | 360 | 90 | 15 |
| MC2 | 125 | 39 | 2 |
| processed cleveland | 303 | 13 | 5 |
| Sensorless drive diagnosis | 58509 | 48 | 11 |
| sonar | 208 | 60 | 2 |
| spambase-10an-nn | 4597 | 57 | 2 |
| wdbc-10an-nn | 569 | 30 | 2 |
| lung-cancer | 32 | 56 | 2 |
| Madelon | 2000 | 500 | 2 |

### B. Experiment Tools

For our experiment, we use 10-fold cross-validation to divide the data set into a training set and a test set. The training set is used for feature selection, and the test set is used for model verification. We use the k-nearest neighbor (KNN) classifier to evaluate the performance of selected features for each feature selection method. The number of neighbors in the KNN was set to five and repeat 10 times to take the average. All programs were implemented using the Python language. In particular, the classification algorithm was implemented using the public package tools in scikit-feature [5], and scikit-learn [18].

### C. Performance Measure

To verify the performance of the proposed algorithm, we use the classification accuracy of the KNN as the performance measure. The accuracy is defined as:

$$acc = \frac{TP + TN}{TP + FP + TN + FN} \qquad (6)$$

True positive (TP): both the actual and predicted classes are positive examples;

False positive (FP): the actual and predicted classes are negative and positive examples, respectively;

False negative (FN): the actual and predicted classes are positive and negative examples, respectively;

True negative (TN): both the actual and predicted classes are negative examples.

## IV. EXPERIMENT RESULTS

In this section, we compared the proposed RFCBF algorithm with FCBF, FCBF#, and FCBFiP, to verify the effectiveness and advantages of RFCBF. All the results reported below were obtained by averaging the values over the 10 runs. The optimal values obtained for each data set are highlighted in bold.

### A. RFCBF Parameter Setting

Since the proposed RFCBF involves three parameters, namely threshold $\delta$, sampling times, and sampling probability, respectively. So that each sample in the data set has the same probability to be sampled, we set the sampling probability value to 0.5. Figure 1 presents the impact of sampling times on the classification accuracy of RFCBF on 12 data sets when the threshold is 0.01 and the sampling times are 5, 10, 15, and 20, respectively. It can be seen from Figure 1 that when the number of sampling times is 20, it achieved the best results on 7 data sets. Figure 2 shows the average accuracy of these four sampling times on 12 data sets. It was discovered that when the number of sampling times is 20, the performance of RFCBF is the best. When the number of samples is 5 and 15, the average performance of RFCBF varies little with the number of samples. Table 2 shows the impact of the performance of RFCBF on the classification accuracy on 12 data sets when the number of sampling times is 20 and the threshold is 0.01, 0.02, 0.03, and 0.04, respectively. It is clear that RFCBF achieved the best results on 6 data sets when the threshold is 0.01. When the threshold is gradually increased, the optimal feature subset selected by RFCBF is empty on some data sets. Table 3 shows the average number of features selected by RFCBF on 12 data sets under these four thresholds.

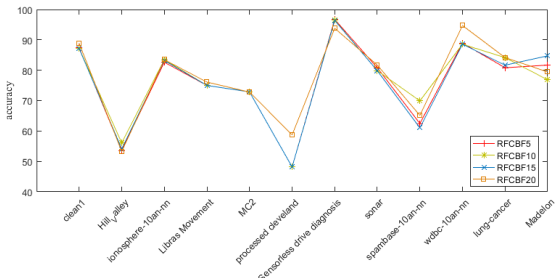

Fig. 1. The effect of sampling times on the accuracy of RFCBF

Therefore, for the setting of RFCBF parameters, we finally set the threshold to 0.01, the sampling probability to 0.5, and the number of sampling times to 20 for comparison experiments between RFCBF and other feature selection methods.

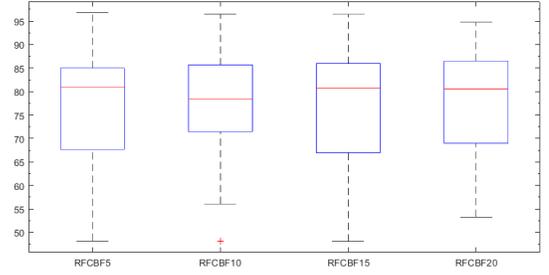

Fig. 2. Average accuracy over data sets with different sampling times

### TABLE 2. THE IMPACT OF THRESHOLD ON RFCBF PERFORMANCE

| Data set | 0.01 | 0.02 | 0.03 | 0.04 |
|---|---|---|---|---|
| clean1 | **88.87** | 86.54 | 84.44 | 83.40 |
| Hill_Valley | 53.28 | **54.60** | NA | NA |
| ionosphere-10an-nn | 83.48 | **83.77** | 81.78 | 81.77 |
| Libras Movement | **76.11** | 75.00 | 75.00 | 75.00 |
| MC2 | 72.82 | **72.88** | 72.88 | **72.88** |
| processed cleveland | **58.72** | 48.17 | 48.17 | 48.17 |
| Sensorless drive diagnosis | 93.82 | **96.41** | **96.41** | **96.41** |
| sonar | 81.71 | 80.81 | **82.21** | 81.76 |
| spambase-10an-nn | **65.13** | NA | NA | NA |
| wdbc-10an-nn | **94.73** | 88.58 | 88.58 | 88.93 |
| lung-cancer | **84.17** | 80.83 | **84.17** | **84.17** |
| Madelon | 79.45 | **81.10** | NA | NA |
| Winners | 6 | 5 | 3 | 3 |

### TABLE 3. AVERAGE NUMBER OF SELECTED FEATURES BY RFCBF

| Data set | 0.01 | 0.02 | 0.03 | 0.04 |
|---|---|---|---|---|
| clean1 | 164.2 | 148.1 | 115.7 | 80.7 |
| Hill_Valley | 67.0 | 55.4 | NA | NA |
| ionosphere-10an-nn | 23.5 | 24.6 | 24.3 | 23.9 |
| Libras Movement | 90.0 | 90.0 | 90.0 | 90.0 |
| MC2 | 37.5 | 37.0 | 37.0 | 36.9 |
| processed cleveland | 13.0 | 12.9 | 12.1 | 10.6 |
| Sensorless drive diagnosis | 14.6 | 14.0 | 12.0 | 12.0 |
| sonar | 59.9 | 58.8 | 53.8 | 42.1 |
| spambase-10an-nn | 2.6 | NA | NA | NA |
| wdbc-10an-nn | 17.9 | 18.6 | 17.7 | 18.1 |
| lung-cancer | 54.2 | 53.1 | 55.1 | 54.9 |
| Madelon | 13.3 | 7.2 | NA | NA |

### B. RFCBF vs. FCBF

Tables 4–5 show the average accuracy of FCBF and RFCBF on 12 data sets, the average number of selected features, and the average runtime required to select features, respectively. In Table 4, compared to FCBF, except for the spambase-10an-nn data set, the proposed RFCBF algorithm yielded the best results in the average accuracy. It is clear that RFCBF achieved the highest accuracy of 94.73 on the wdbc-10an-nn data set and the lowest accuracy of 53.28 on the Hill_Valley data set. In Table 5, from the overall point of view of the 12 data sets, in spite of the average number of features selected by RFCBF is more than that of FCBF, the average runtime is much less than that of FCBF. We can see in Table 5 that for FCBF, the longest runtime data set is Sensorless drive diagnosis, which takes 2318.5835s, and the shortest is lung-cancer, which takes 0.1944 s. For RFCBF, the longest runtime data set is clean1, which takes 790.8440 s, and the shortest is processed cleveland, which takes 1.0641s.

Based on the above analysis, from the perspective of classification accuracy and average running time, RFCBF has significant advantages over FCBF.

TABLE 4. THE AVERAGE ACCURACY OF KNN ON SELECTED FEATURES FOR FCBF AND RFCBF

| Data set | FCBF | RFCBF |
|---|---|---|
| clean1 | 61.99 | **88.87** |
| Hill_Valley | 47.50 | **53.28** |
| ionosphere-10an-nn | 79.48 | **83.48** |
| Libras Movement | 11.67 | **76.11** |
| MC2 | 56.60 | **72.82** |
| processed cleveland | 51.53 | **58.72** |
| Sensorless drive diagnosis | 11.34 | **93.82** |
| sonar | 52.48 | **81.71** |
| spambase-10an-nn | **72.96** | 65.13 |
| wdbc-10an-nn | 83.31 | **94.73** |
| lung-cancer | 83.33 | **84.17** |
| Madelon | 51.75 | **79.45** |
| Mean | 55.33 | **77.69** |

TABLE 5. THE AVERAGE NUMBER OF SELECTED FEATURES AND THE AVERAGE RUNTIME(IN S) FOR FCBF AND RFCBF

| Data set | selected features | | runtime | |
|---|---|---|---|---|
| | FCBF | RFCBF | FCBF | RFCBF |
| clean1 | 1.0 | 164.2 | 14.8534 | 790.8440 |
| Hill_Valley | 1.0 | 67.0 | 17.4515 | 129.4540 |
| ionosphere-10an-nn | 2.0 | 23.5 | 2.0282 | 7.1047 |
| Libras Movement | 1.0 | 90.0 | 7.3621 | 161.9676 |
| MC2 | 1.0 | 37.5 | 1.2796 | 15.5445 |
| processed cleveland | 1.1 | 13.0 | 0.4732 | 1.0641 |
| Sensorless drive diagnosis | 1.0 | 14.6 | 2318.5835 | 452.1245 |
| sonar | 1.0 | 59.9 | 5.3528 | 32.5704 |
| spambase-10an-nn | 1.0 | 2.6 | 29.4513 | 1.6766 |
| wdbc-10an-nn | 1.0 | 17.9 | 4.1738 | 6.1953 |
| lung-cancer | 5.7 | 54.2 | 0.1944 | 26.2767 |
| Madelon | 1.0 | 13.3 | 34.2813 | 98.7034 |
| Mean | 1.5 | 46.5 | 202.9571 | 143.6272 |

## C. Compared with other state-of-the-art feature selection methods

In the above section, we confirmed that the proposed algorithm RFCBF has a huge advantage over FCBF in classification accuracy and runtime, especially in improving classification accuracy. To further verify the efficiency of RFCBF, we compared the two feature selection methods FCBF# and FCBFiP, which are extended by FCBF. To compare with RFCBF more fairly, we take the number of features selected by RFCBF as the criterion. We select the same number of features ranked at the top of each algorithm as the feature number of the algorithm. Since the number of features selected by RFCBF in the 12 data sets contains decimals, we set the number of features to be selected by FCBF# and FCBFiP by performing dead-round rounding operations.

TABLE 6. THE AVERAGE ACCURACY OF KNN ON SELECTED FEATURES BY RFCBF, FCBF#, AND FCBFIP

| Data set | RFCBF | FCBF# | FCBFiP |
|---|---|---|---|
| clean1 | **88.87** | 88.67 | 88.67 |
| Hill_Valley | **53.28** | 52.46 | 52.46 |
| ionosphere-10an-nn | 83.48 | 80.90 | **83.76** |
| Libras Movement | **76.11** | **76.11** | **76.11** |
| MC2 | **72.82** | 70.45 | 71.28 |
| processed cleveland | **58.72** | **58.72** | **58.72** |
| Sensorless drive diagnosis | **93.82** | 57.58 | 93.75 |
| sonar | **81.71** | **81.71** | **81.71** |
| spambase-10an-nn | 65.13 | **79.66** | 58.30 |
| wdbc-10an-nn | **94.73** | 94.56 | 91.75 |
| lung-cancer | **84.17** | 75.00 | 71.67 |
| Madelon | 79.45 | **81.20** | 52.00 |
| Mean | **77.69** | 74.75 | 73.35 |

TABLE 7. COMPARISON OF AVERAGE RUNTIME (IN S) OF FEATURE SELECTION FOR RFCBF, FCBF#, AND FCBFIP

| Data set | RFCBF | FCBF# | FCBFiP |
|---|---|---|---|
| clean1 | 790.8440 | 7.5639 | 316.1262 |
| Hill_Valley | 129.4540 | 12.8314 | 298.6058 |
| ionosphere-10an-nn | 7.1047 | 2.1589 | 11.3423 |
| Libras Movement | 161.9676 | 4.2863 | 105.8104 |
| MC2 | 15.5445 | 0.5886 | 3.6804 |
| processed cleveland | 1.0641 | 0.2577 | 0.4432 |
| Sensorless drive diagnosis | 452.1245 | 1804.1385 | 2342.3398 |
| sonar | 32.5704 | 2.3711 | 35.6775 |
| spambase-10an-nn | 1.6766 | 23.732 | 589.2106 |
| wdbc-10an-nn | 6.1953 | 2.4422 | 16.4114 |
| lung-cancer | 26.2767 | 0.1019 | 0.9358 |
| Madelon | 98.7034 | 116.1276 | 37.1376 |
| Mean | 143.6272 | 164.7167 | 313.1434 |

Table 6 shows the average accuracy on 12 datasets for the KNN classifier based on RFCBF, FCBF#, and FCBFiP, respectively. The RFCBF algorithm achieved competitive results on the 9 datasets. FCBF# and FCBFiP achieved the best results on 5 data sets and 4 data sets, respectively. On the Libras Movement and sonar data sets, the classification accuracy of the three feature selection methods is the same. From the overall average classification accuracy of the 12 data sets, RFCBF ranked first with an average accuracy of 77.69, followed by FCBF# (74.75), and finally FCBFiP (73.35). This also further shows that the proposed RFCBF algorithm has advantages in improving classification accuracy.

Table 7 lists the average runtime of feature selection for the RFCBF, FCBF#, and FCBFiP on each data set. The three methods can be directly compared by verifying the mean values in the last row in Table 7. As shown, RFCBF indicated the shortest execution time; FCBFiP required the longest time; FCBF# required a longer time than RFCBF but less time than FCBFiP. This also verifies that RFCBF has a huge advantage in feature selection time-consuming.

## V. CONCLUSION AND FUTURE WORK

In this paper, we proposed an extension of FCBF by combining resampling technology, called RFCBF. Compared with FCBF, RFCBF has stronger robust and adaptive capacity. From the overall perspective of the experiment on 12 benchmark datasets, the proposed RFCBF algorithm has a huge advantage not only in classification accuracy but also in runtime compared with FCBF, FCBFiP, and FCBF#.

In addition to the threshold $\delta$, the proposed RFCBF algorithm also involves parameters of sampling probability and sampling times. In future work, we will test more data sets to explore the effect of sampling probability and sampling times.

Acknowledgements

Research on this work was partially supported by the grants from Jiangxi Education Department of China (No. GJJ201917, No. GJJ170995), and National Science Foundation of China (No.61562061).